\documentclass[11pt]{article}
\usepackage{sectsty}
\usepackage[utf8]{inputenc}
\usepackage[margin=1cm]{caption}
\usepackage[
    left=1.5cm,
    right=1.5cm,
]{geometry}
\usepackage{amsmath}

\usepackage{graphicx} 
\usepackage{float}

\usepackage{booktabs} 
\usepackage{array} 
\usepackage{paralist} 
\usepackage{verbatim} 
\usepackage{subfig} 

\usepackage[lined,vlined]{algorithm2e}

\usepackage{fancyhdr}
\usepackage{epstopdf} 
\usepackage{graphicx}          
\usepackage{amsmath} 
\usepackage{amssymb}  
\usepackage{textcomp}

\usepackage{gensymb}

\usepackage{appendix}

\SetKwBlock{Lock}{lock}{release}
\SetAlFnt{\small\sf}

\usepackage{fancyhdr} 
\allsectionsfont{\sffamily\mdseries\upshape} 

\usepackage[nottoc,notlof,notlot]{tocbibind} 
\usepackage[titles,subfigure]{tocloft} 

\newtheorem{definition}{Definition}
\newtheorem{theorem}{Theorem}
\usepackage{etoolbox}
\usepackage{pgf}
\usepackage{xcolor}

\usepackage{highlight}

\title{Long-Range Route-planning for Autonomous Vehicles in the Polar Oceans}
\author{Maria Fox$^1$, Michael Meredith$^1$, J. Alexander Brearley$^1$, Dan Jones$^1$ and Derek Long$^2$ }
\date{%
    $^1$British Antarctic Survey\\%
    $^2$Schlumberger Cambridge Research\\[2ex]%
}

\begin{document}
\maketitle

\begin{abstract}
    {There is an increasing demand for piloted autonomous underwater vehicles (AUVs) to operate in polar ice conditions. At present, AUVs are deployed from ships and directly human-piloted in these regions, entailing a high carbon cost and limiting the scope of operations. A key requirement for long-term autonomous missions is a long-range route planning capability that is aware of the changing ice conditions. In this paper we address the problem of automating long-range route-planning for AUVs operating in the Southern Ocean. We present the route-planning method and results showing that efficient, ice-avoiding, long-distance traverses can be planned.
    
    \vspace{\baselineskip}
    \noindent This paper is under review\footnote{This work has been submitted to the Journal of Atmospheric and Oceanic Technology. Copyright in this work may be transferred without further notice.}.
    }
\end{abstract}

\section{Introduction}
\label{sec:intro}

In the last decade, the rising demand for oceanographic data, to inform diverse contemporary issues such as climate change, ecosystems services and marine conservation, has spurred the development of autonomous platforms for sustained acquisition of data. To extend the capability of existing networks such as Argo \cite{Riser2016}, there is considerable interest in increasing the coordinated use of piloted autonomous underwater vehicles (AUVs)~\cite{Testor2019} in the hostile conditions of the polar oceans. Such vehicles, including low-power underwater gliders, offer distinct advantages to ocean observing networks.  They can be piloted remotely and hence can conduct targeted campaigns in specific areas, with tasking of the vehicles changing during missions in response to evolving environmental conditions or user requirements.  

At present, such autonomous underwater vehicles are commonly used individually or in small clusters, with direct human piloting of individual vehicles from land. There is strong interest in progressing capability to the point where these vehicles can be deployed and controlled fully autonomously, in complex environmental conditions and over long periods of time. For this capability to be achieved in the polar oceans, the path-planning methods used must take account of the changing ice conditions. In this paper we address the problem of automating long-range route-planning for AUVs operating in the Southern Ocean. 

\section{Background: Automated Long-distance Route-planning}
We address the problem of finding a navigable path between two locations, whilst respecting a collection of spatial and temporal environment constraints, and optimising an objective function.

There is a large body of work in automated path planning for robotic vehicles, including underwater vehicles. The most common methods use grid-based discretisations or continuous numerical approaches. Grid-based methods discretise the environment into a graph of locations and then use optimisation methods to find paths from the start node to the destination. Approaches to the problem of finding shortest paths in grids use heuristic search~\cite{Carroll1992} or greedy methods such as Dijkstra's algorithm~\cite{dijk,Sen2015}. Using a graph-based representation, in which the environment is gridded at some chosen resolution, edges can be weighted to model environmental impacts on travel-time, such as the presence of strong surface currents. A shortcoming of grid-based approaches is that all paths follow the edges of the pre-defined grid, between vertices that are the centres and corners of the cells. This results in so-called Manhattan paths, consisting of straight horizontal, diagonal and vertical lines between cell centres, which are often far from optimal depending on the grid resolution. Further increasing grid resolution to better approximate optimal solutions is often prohibitively computationally expensive. A further difficulty arises when environmental constraints are time-dependent. These cannot easily be combined into edge weights, because they change over time whilst the edge weights are constant. The 2D grid used by Dijkstra's algorithm must be extended to a 3rd dimension to model temporal variability, which introduces extra complexity.

An alternative, cell-free, numerical modelling approach is to use the principle of wave propagation to identify the points in an inhomogeneous space that can be reached in equal time (or energy expenditure) from the location of the vehicle. The approach avoids discretisation of the space at the outset, but faces similar problems of efficiency and optimality of solutions that depend on the number of points identified on each wavefront. Furthermore the need for a temporal dimension must still be addressed when conditions are time-dependent.

Both the grid-based and wave propagation methods have been used in long distance route planning for both AUVs and ships~\cite{bijlsma1975,Orioke2019,Petres2007,Li2021,Topaj2019}. Long-range AUV operations in non-polar conditions are well-established: the Tethys Long-Range AUV has successfully completed many science surveys consisting of long traverses in the Pacific Ocean, of up to 3000km~\cite{Bellingham2012}, using a path-planning method based on Dijkstra's algorithm~\cite{Orioke2019} with the goal of optimising energy efficiency. 

\begin{figure}
\centerline {\includegraphics[width=8cm]{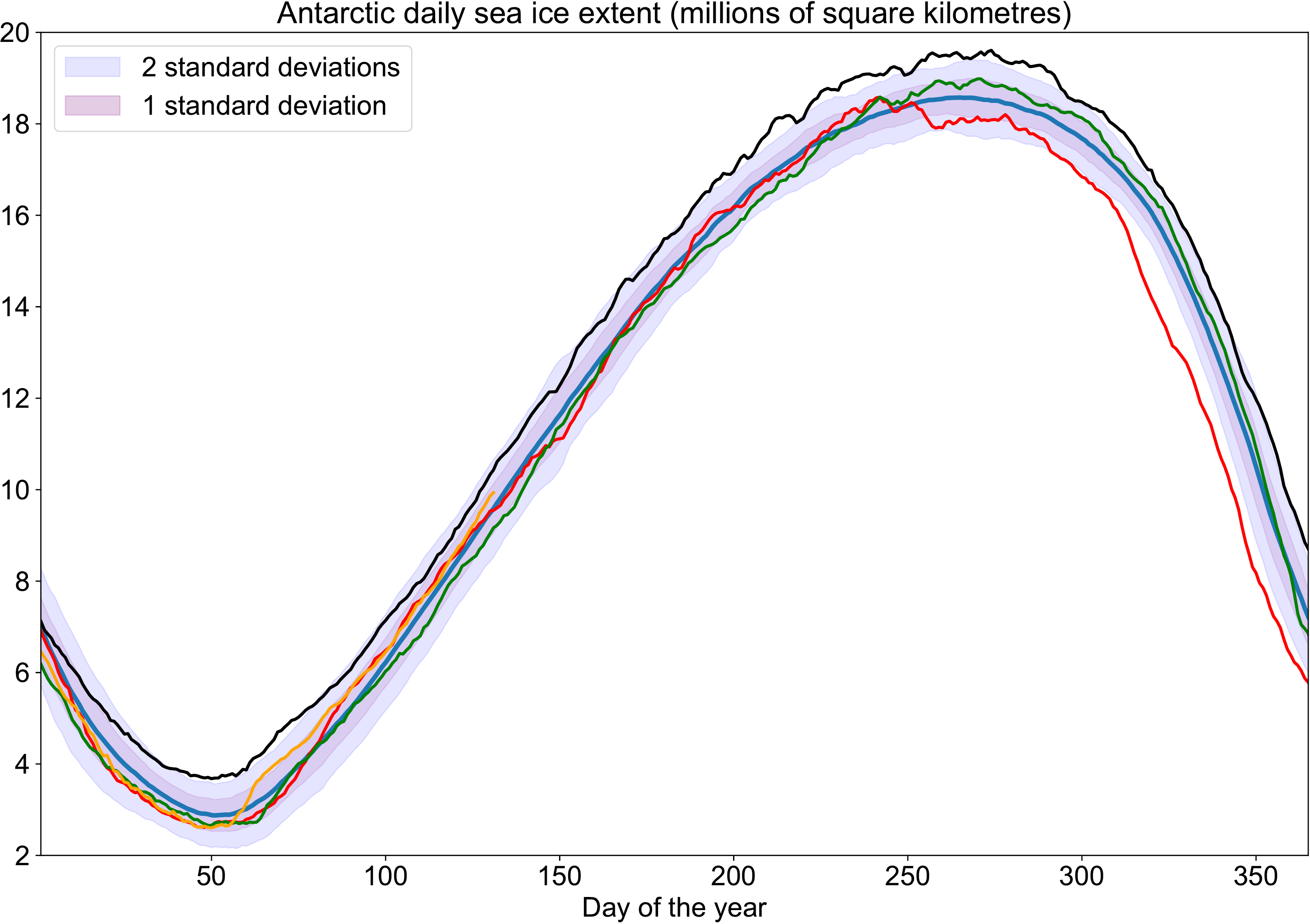}}
\caption{The seasonal pattern of sea ice extent in the Antarctic. The blue line is the mean sea ice extent over the years 1978-2021. Two standard deviations are shown. The black line is 2013, a year of maximal ice extent. The green line is 2010. The red line is 2016, a minimal year. The orange line is 2021.}
\label{fig:seaiceextent}
\end{figure}

Path-planning methods for underwater vehicles have thus far not considered long-distance route planning in environments containing sea ice or icebergs. Sea ice extent in the polar oceans varies month-by-month and year-by-year, presenting many dangers and challenges to autonomous platforms. The historical pattern of ice extent growth and retreat for the Antarctic is shown in Figure~\ref{fig:seaiceextent}. Path-planning in these regions must take these seasonal patterns into account as well as the behaviours of the ice in areas characterised by perennial sea ice, such as the Weddell Sea.

\section{Long-distance Route-planning in Currents and Ice}
Our objective is to automate long-distance route-planning for AUVs in the polar oceans. AUV pilots typically focus on keeping well clear of the ice edge, unless the AUV in question has specific ice-coping capabilities, which most commercially available products do not. Our autonomously planned routes will likewise avoid dangerous ice conditions whilst exploiting opportunities to travel as close to the ice edge as is possible without compromising the safety of the vehicle.

In the approach described in this paper we grid the environment and use Dijkstra's algorithm to find routes that minimise travel time. We address the limitations of Dijkstra's algorithm, noted above, in three ways. First, we support a non-uniform gridding so that cells can be refined at the ice edge. Second, we mitigate the Manhattan path problem by using Newton's method to find the optimal crossing point on the edge between two adjacent cells being passed through on a route. This effectively allows an infinite set of vertices along every edge, enabling substantial smoothing of routes without significant computational expense (Newton's method quickly converges to the optimal crossing point on the edge). Finally, we add a temporal dimension by discretising time into monthly steps and recomputing the gridding and optimisation steps in each month. This combination leads to a computationally inexpensive 3D (two spatial dimensions and one temporal dimension) shortest path method that gives good estimates of the costs implied by AUV journeys between distant waypoints. These can be readily recomputed during operation, in response to unexpected changes in conditions, giving an adaptive route-planning capability. 

In this paper we focus on refining the grid along the ice edge, but the refinement principle can also be used to refine the grid in other contexts, for example: in the presence of land, ocean fronts, eddies, sharp bathymetric gradients and so on.  A principle of {\em homogeneity} is defined, which determines the subdivision process. Cell sub-division is performed whenever a cell (of any size) is determined to be inhomogeneous with respect to a specific characteristic of interest.  Routes can then be planned using Newton's method to find paths between adjacent cells and Dijkstra's algorithm to find optimal routes between waypoints through the non-uniform grid. We refer to this combination of Dijkstra's and Newton's methods as the {\em 3D dynamic shortest path} (3D-DSP) method.

The cost function may be any metric objective, such as travel time, distance, power demands, fuel consumption or others, or indeed some combination of metrics. In the work presented here, we optimise travel time, as this is an important consideration for route-planning in dynamic conditions, and can also be interpreted as a simple proxy for energy consumption. The 3D-DSP method can be used in the spatially and temporally varying conditions of the polar oceans, to calculate the fastest paths and travel-times, for autonomous vehicles such as gliders, powered AUVs and other controllable vehicles that may form part of a fleet. 
\section{Environmental Setting, Data and Methods}
\label{sec:methods}
In this work, we use the Southern Ocean as an example domain. The region of interest extends from 130$^\circ$W to 30$^\circ$E, and 40$^\circ$S to 80$^\circ$S (i.e. the Atlantic sector of the Southern Ocean including the Weddell Sea). This region, an area of more than 2.8 million km$^2$, was chosen since it contains all of the natural hazards present in any area of the global ocean, plus additional factors such as sea ice.
The model of currents and ice extent is built using observationally constrained and physically consistent current data and sea ice concentration data from the Biogeochemical Southern Ocean State Estimate (B-SOSE) reanalysis product~\cite{Mazloff2010,Verdy2017}. 

\begin{figure}
\centerline {\includegraphics[width=8cm]{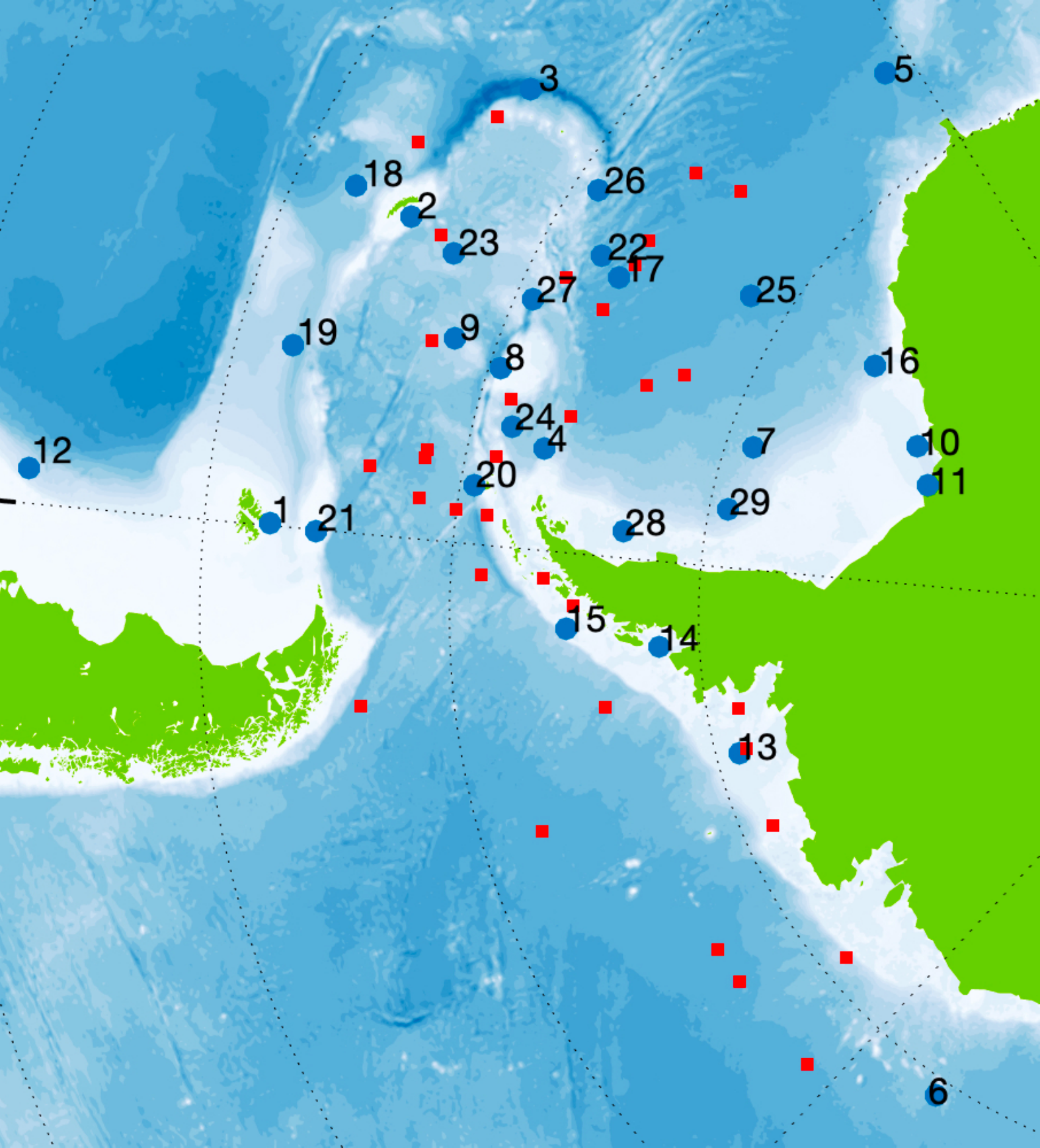}}
\caption{Map showing the waypoints (blue numbered dots) of scientific interest listed and named in Appendix B. The unlabelled red points are scattered additional waypoints.}
\label{fig:mapofwaypo}
\end{figure}

\subsection{Representation of the ocean environment}
\label{sec:oceanenv}

The model of the environment is described in three stages. First, we describe an initial procedure for coarsely gridding the region of interest shown in Figure~\ref{fig:mapofwaypo}. The second stage is to introduce a temporal abstraction on top of the spatial abstraction of the initial coarse grid. This allows the representation of features that change over time, such as surface ice cover. The third stage is to identify areas within the spatio-temporal abstraction where grid refinement is needed in order to capture a description of the local sea ice conditions that is adequate for safe route-planning. 

\subsubsection{Initial coarse grid}
\label{spatial}
\begin{figure}
\centerline {\includegraphics[width=5cm]{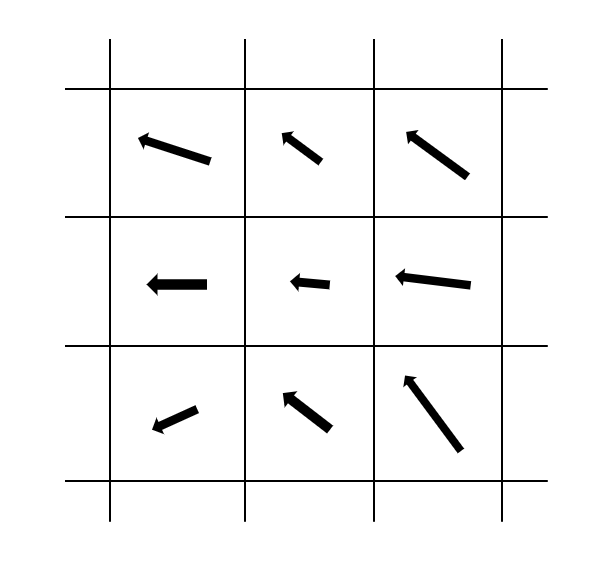}}
\caption{Part of the coarse grid abstraction of the vector field. Each cell contains one vector which is an average of the currents in that cell.}
\label{fig-cellboxes}
\end{figure}

The B-SOSE data set used for this example contains current vectors each at a specified  coordinate in the Southern Ocean. We start by constructing an abstraction of the field into a 2-dimensional grid of cells, each one associated with an array containing the vectors belonging to that cell. To characterise the cell, we compute its average current vector.  Each cell contains a centre point which is used for route-planning. Figure~\ref{fig-cellboxes} shows an arrangement of cells with the averaged current vector in each cell.

Cell dimensions of 5$^\circ$ longitude by 2.5$^\circ$ latitude are use (this is configurable). Given that the input file contains vectors at intervals of one sixth of a degree, the maximum number of vectors that a 5 $\times$ 2.5 degree cell can contain is 450. If it contains many fewer than this, then the cell is likely to lie at least partially on land and may therefore not be navigable. In the coarse grid a cell is classified as land if it contains fewer than 25\% of the maximum number of vectors.

\subsubsection{Spatio-temporal abstraction}
For each year in our data set, we extend the spatial abstraction described in \ref{spatial} to take into account the presence and extent of sea ice in the same region. The ice cover each year has a strong seasonal pattern, as is clear in Figure~\ref{fig:seaiceextent}, although this pattern has been changing over the last decade. For route-planning, a monthly abstraction has been chosen. This is detailed enough to enable the route planner to take into account the seasonal pattern without becoming overwhelmed by detail on other timescales.

Each cell in the initial coarse grid, as well as recording the average current vector, now records a list of 12 monthly abstractions.  A monthly abstraction contains the mean and variance of the sea ice area at that location over the corresponding month. This extends the 2-dimensional representation into a 3-dimensional one in which there are twelve layers, each a grid representing the region of interest in a given month.

A simple conservative threshold ice cover is used for determining that a cell is {\em ice-locked} and therefore inaccessible. Whether or not a cell is ice-locked changes from month to month. This means that some cells are navigable during certain months and not others, depending on the year being modelled.

\subsubsection{Refining the grid}
\label{refgrid}

A cell of size 5 $\times$ 2.5 degrees covers an area of more than 154,000km$^2$ at the Equator. Cells closer to the pole are much smaller than this, for example a cell at 60$^{\circ}$S covers about 77,000 km$^2$. This is still extremely large and adequate only for describing huge, deep, invariant regions of open sea where the variance in current is small and there is no ice present. A non-uniform grid is proposed so that, in complex regions such as the Weddell Sea, cells of the grid are recursively subdivided.

We assume that refinement of the grid is based on the presence of data that characterises some property or properties of each of the cells. Where that data can be adequately characterised by a single model within a cell, we do not need to refine it further, but in the cases where the data cannot be appropriately characterised, the cell is decomposed. The spatial decomposition is based on the use of a quad-tree~\cite{quadtrees} (widely used in representation of 2-dimensional data, such as images), refining a cell into quarters. The data in the cell is then reallocated amongst the subdivisions and the process of subdivision applied recursively to each of the quarters. We refer to the requirement that must be satisfied in order to avoid further need for subdivision as a {\em homogeneity} condition -- homogeneous data needs no further division, while inhomogeneous data requires that the containing cell be split. This process can be continued arbitrarily far, but, in practice, once the data has been split a number of times, the data content in each cell can become too sparse to make further subdivision practical. Therefore, the definition of homogeneity of the data typically includes a threshold on the quantity of data.

Definition~\ref{homog} shows how we define a homogeneity condition, $H$, for our context. 
\begin{definition}\label{homog}  A cell is considered {\em homogeneous} if, for every data set $D$ recorded in the cell, a statistical measure, $g(D)$, is below the homogeneity threshold.
\end{definition}
The splitting process is implemented using Algorithm~\ref{alg1}. The input to the algorithm is:
\begin{itemize}
    \item a cell, $C$, with bounds $a$ and $b$ (latitudes) in the $x$-dimension, and $c$ and $d$ (longitudes) in the $y$ dimension;
    \item a data set of values $D$, with each datum associated with a point in $C$ by the function $f$;
    \item a boolean function, $H$, called the {\em homogeneity condition};
\end{itemize} 
 $H$ is a boolean function that decides whether or not a cell should be split. The statistical measure required to compute $H$, and the associated threshold, will vary depending on the feature being considered. A detailed example is given in Section~\ref{sec:nonhomog}, in which $H$ returns {\em true} (the cell is homogeneous and should not be split) if the data set of ice concentrations in the cell falls outside a specified range. If the homogeneity condition is not satisfied, then the cell is split.

\begin{algorithm}[H]
\bf{Algorithm: Split(C,D,f,H)}\\
\KwData{A 2-dimensional cell, $C = \{(x,y) \,|\, a \leq x < b, c \leq y < d\}$, a dataset, $D$, a function, $f : D \rightarrow C$, associating data in $D$ with points in $C$ and a homogeneity condition, $H : \mathcal{P}(D) \rightarrow Boolean$ on subsets of data in $D$}
\KwResult{A partition of $C$ into cells $\mathcal{C}$}
\Begin{
$\mathcal{C} \longleftarrow \emptyset$\;
\If{H(D)} {
add $C$ to $\mathcal{C}$\;
\Return{$\mathcal{C}$}\;
}
\Else{
$C_1 \longleftarrow \{(x,y)\,|\,(x,y) \in C, x < (a+b)/2, y < (c+d)/2\}$\;
$C_2 \longleftarrow \{(x,y)\,|\,(x,y) \in C, x < (a+b)/2, y \geq (c+d)/2\}$\;
$C_3 \longleftarrow \{(x,y)\,|\,(x,y) \in C, x \geq (a+b)/2, y < (c+d)/2\}$\;
$C_4 \longleftarrow \{(x,y)\,|\,(x,y) \in C, x \geq (a+b)/2, y \geq (c+d)/2\}$\;
\For{i = 1 \KwTo 4}
{
    $D_i \longleftarrow \{ d \,|\, d \in D, f(d) \in C_i\}$\;
    $\mathcal{C} \longleftarrow \mathcal{C} \cup Split(C_i,D_i,f,H)$\;
}
\Return{$\mathcal{C}$}\;
}
}

\caption{Subdivide Cell, $C$ into partitions according to homogeneity condition, $H$, applied to content data $D$.}\label{alg1}
\end{algorithm}

Note that Algorithm~\ref{alg1} is recursive, allowing each subcell to be independently further subdivided. The condition, $H$, checks both that there is sufficient data in the subcell to support continued subdivision and also that the data satisfies the required statistical property.

The construction of a non-uniform grid based on the homogeneity principle allows us to combat the deficiencies of the coarse grid, while retaining the benefits of a sparse structure where possible.

\subsubsection{Inhomogeneity in the local ice concentration}
\label{sec:nonhomog}

We define a range, from a lower bound $lb$ to an upper bound $ub$, and a threshold, $t$. Then, a cell month is considered inhomogeneous if between $lb$ and $ub$ of the ice measurements in that cell month are at $t$\% or higher.  If the proportion of ice in the cell month above the $t$\% concentration is below $lb$\%, we consider the cell month to be homogeneous open water: such a cell can be navigated through in that month. At the other end of the range, if the proportion is greater than $ub$\%, then the cell month is considered homogeneous ice. If the proportion is between these bounds, then the cell is inhomogeneous and must be split so that the homogeneous sub-cells can be found.

Given a cell month, $C$, and a data set of sea ice concentrations, $D$, associated with $C$, we define a statistical measure, $g(D)$, as follows:

\begin{equation}
\label{eqn:eq1}
g(D) = (lb - x)(x - ub),
\,\,\, {\mbox where } \,\,x = \frac{|\{d \in D\, |\, ice(d) > t\}|}{|D|}
\end{equation}

The function $ice(d)$ returns the sea ice concentration (as a percentage) associated with the datum $d$. The homogeneity condition, $H(D)$, which decides whether the cell should be split, is $g(D) > 0$, which is true whenever the proportion of ice above the threshold lies inside the specified range. The sub-splitting method quickly focuses on the parts of the cell where the high percentages of ice reside in the cell and month being considered, enabling the vehicle to traverse the ice edge more nimbly than is possible with the uniform grid abstraction. 

The degree of refinement of the grid impacts on waypoint accessibility and degree of exposure to risk, therefore we consider two different homogeneity conditions: Weak homogeneity:  $lb = 15$, $ub = 70$, and Strong homogeneity:  $lb = 5$, $ub = 90$. A $4\%$ threshold is used in both cases, in order to focus on the impact of the size of the range. 

The intuition is that a homogeneity condition is weak if $g(D)>0$ is hard to satisfy (the range is small), as this results in a lower level of discrimination between different ice conditions. It is strong if, by contrast, $g(D)>0$ is easy to satisfy (the range is large), as this encourages sub-division and hence differentiation. When $lb = ub$, the homogeneity condition is always true and the grid is never refined. 

Since subdivision divides a cell into quarters, only 8 subdivisions are required to reduce a 5~$\times$~2.5 degree cell, of area 154,000~km$^2$ at the Equator, to an array of cells the smallest of which is 2~km$^2$ in area. In practice, the number of subdivisions possible is limited by the quantity of data available in the resulting cells. Given that the B-SOSE data set supplies measurements at one sixth of a degree, after 3 subdivisions a subcell contains only 7 daily data points and, hence, 210 points for any given month. If further splitting is performed, the ability to perform meaningful statistical analyses is lost. We therefore restrict the depth of subdivision to 3 for the experiments reported in this paper.

The degree of refinement of the grid varies between layers, reflecting the fact that the sea ice area changes from month to month. Additionally, the sea ice changes between years, so the 3-dimensional grids representing different years may have very different levels of refinement in each of the layers. 

\subsection{The 3D-DSP Method}
\label{newtopt}
The grid abstraction represents the region as a graph, 
\begin{equation}
G = \langle V,E\rangle
\end{equation} 
where $V$ is the set of vertices (cell centre points) and $E$ is the set of edges (lines connecting adjacent pairs in $V$). As part of the problem specification we are given a special set of  coordinates called the waypoints set (a complete list is provided in  Appendix B). These are identified locations with the special significance that the routes to be planned will start and end at them. Every point in the waypoints set exists inside the cell whose central point is closest to the waypoint  coordinate. 

The full 3D-DSP method involves two stages. First, the shortest paths are computed from each source waypoint to all destination waypoints {\em within} each month (the planning month), and hence each layer, of the 3D grid. The planning month is therefore fixed for each path, even though paths may be longer than 1 month in duration. 

The travel time from one waypoint (the source) to another (the destination), within the layer characterising the planning month, can be found by using our 3D-DSP method to compute the travel time from the centre of the cell containing the source, to the centre of the cell containing the destination, and adding the travel times required to get from the source and destination to and from their respective centre points. The entire path is constructed within the horizontal layer. The resulting route and travel-time is stored in a structure called the {\em path-book}, indexed by the planning month. This first stage is called the path-book construction stage.

Given a suitably dense set of waypoints, many of the paths in the path-book are of less than one month duration. However, the path-book construction stage does not limit routes to be no more than one month long. Some routes can be completed beyond the outer limits of the ice edge and therefore are not affected by the changing sea ice extent. In cases where a long route does visit waypoints in ice-prone areas, planning within a layer results in unexecutable routes because the ice conditions relevant to later parts of the routes are different from those modelled in the planning month. Even these routes might have long sections that are independent of the ice. 

The second stage runs Dijkstra's algorithm, over a meta-level graph, to find optimal multi-month routes by considering compositions of the routes in the path-book. The composition process steps vertically between layers in the 3-dimensional grid, allowing different parts of the routes to be planned under different monthly abstractions. Sometimes, several layers must be traversed in one vertical step, corresponding to time spent waiting for a waypoint to become accessible. We impose the constraint that waiting time cannot be spent in ice-bound locations. This second stage is called the path-composition stage and is described in detail in section~\ref{sec:compos}.

\subsubsection{Route-planning}
\label{newtderiv}
We begin by describing how the horizontal routes comprising the path-book are constructed using Dijkstra's and Newton's methods.
\begin{definition} \label{defn-adj} Two cells, A and B, are {\em adjacent} if the straight line connecting the centre of A to the centre of B passes through no other cell.
\end{definition}

A key question is how best to pass between adjacent cells in a grid defined by a given homogeneity configuration. When all cells are the same size, a given cell can have at most 8 adjacent cells. In the refined grid, a cell can have as many as 36 adjacent cells when the depth of subdivisions is limited to 3 (8 on each side plus 4 corner cells). Each subdivision approximately doubles the number of adjacency relationships that must be considered.

The path between two adjacent cells is found by optimising travel in the cell array representation of the vector field. The non-diagonal cases to be considered are adjacent arrangements where the cell being travelled to is at the left, right, above or below the cell being travelled from. All these arrangements can be normalised into an adjacent pair where the task is to go from the left centre to the right centre. This requires rotation of the arrangement and consequent interpretation of the $x$ and $y$ axes and the current vectors.  After normalisation, the path between the adjacent cells goes through an intermediate point on the edge between them, chosen to optimise the travel time between their centre points. This is of course not always the horizontal line, because of the currents present in the two cells. We use Newton's method to optimise the choice of this point, which we call $yval$.

\begin{figure}
\centerline {\includegraphics[width=7cm]{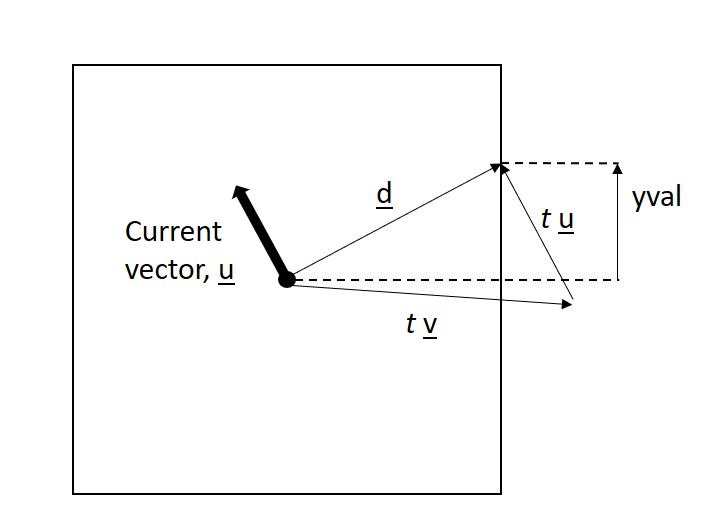}}
\caption{The vector diagram showing how the time taken to travel through the left-hand cell is calculated.}
\label{figvecs}
\end{figure}

In a constant current, the fastest route between two points will be the straight line between them. To achieve the straight line path the vehicle must maintain a constant net velocity along that line, which will be the sum of the current velocity and the velocity the vehicle is maintaining relative to the water. If the vehicle crosses the central boundary between cells at $yval$ (measured relative to the central crossing point), then the time, $t$, it takes to travel from the centre of the cell to the crossing point on the boundary satisfies: 
\begin{equation}
\label{eqn:eqn3}
t (\vec{u} + \vec{v}) = \vec{d} 
\end{equation}
where $\vec{u}$ is the current vector in the cell, $\vec{v}$ is the velocity of the vehicle and $\vec{d}$ is the vector from the centre of the cell to the crossing point on the boundary. The speed of the vehicle is constant on this path and has magnitude $s = |\vec{v}|$. Rearranging Equation~\ref{eqn:eqn3}, we have: \begin{equation}
\label{eqn:eqn4}
t \vec{v} = \vec{d} - t \vec{u}
\end{equation}
the square of which is: 
\begin{equation}
\label{eqn:eqn5}
s^2t^2 = |u|^2t^2 - 2Dt + |d|^2
\end{equation}
where $D = \vec{u}.\vec{d}$. This equation can be solved for $t$: 
\begin{equation}
\label{eqn:eqn6}
t =\frac{ \sqrt{D^2 + |\vec{d}|^2(s^2 - |\vec{u}|^2)} - D}{s^2 - |\vec{u}|^2} 
\end{equation}
with a special case when $s^2 = |\vec{u}|^2$ which is when the speed of the current matches the speed of the vehicle. In this case, $t = \frac{|\vec{d}|^2}{D}$, being undefined when $D = 0$ (the case in which the current is orthogonal to the intended direction of travel). If $D$ is negative, this latter case is degenerate, with the vehicle not fast enough to overcome the current in the intended direction. The solution for $t$ defines the function $ttl(yval)$ for a given current vector, cell size and speed. The relevant vectors are shown in Figure~\ref{figvecs}.

Our function is then the sum of the two travel times, $ttl(yval) + ttr(yval)$ , which we want to minimise by choosing the best $yval$ possible. The iteration for Newton's method is then: 
\begin{equation} 
\label{newton}
yval_{k+1} =  yval_k - \frac{ttl'(yval_k)+ttr'(yval_k)}{ttr''(yval_k)+ttr''(yval_k)}
\end{equation}

When the adjacent pair is diagonally arranged, the travel time between them is initialised to be the straight line between their centres through the corner point connecting the two cells. 

In the refined parts of the grid, the fastest route between two adjacent cells of different sizes will be the line through a $yval$ that is displaced up or down by an amount depending on the number of subdivisions performed. The fact that the fastest path will always be a straight line that passes through the edge of the smaller cell is proven in Theorem~\ref{theorem} in Appendix A.

The details of the modifications to Equations~\ref{eqn:eqn3}-~\ref{newton}, required to manage adjacent cells of different sizes, are given in Appendix C. The method can be called lazily to compute only the travel times required for the specific Dijkstra computation. When the required $yvals$ have been determined, Dijkstra's algorithm can select the best sequence of cells to pass through en route between two waypoints.

As well as responding to currents, planned routes should expose the vehicle to as little risk as possible. The formal definition of risk is as the product of the probability of an event occurring and the severity of the event. The probability of an event occurring depends on a model of the interaction between the vehicle and ice at different concentrations over time, and deriving this model is beyond the scope of this paper. We therefore leave the development of a full risk model for future work, and instead consider the simple concepts of {\em risk avoidance} and {\em risk exposure}. Risk avoidance is achieved during route-planning, by the route planner choosing not to enter cells with a mean ice concentration above a given threshold, while risk exposure is taken to be the time for which a vehicle is exposed to mean ice concentrations above some threshold during simulation of a route. As well as using the effect of the current to determine which adjacent cell to visit, Dijkstra's algorithm compares the mean ice concentration of each cell, according to the monthly abstraction in which the route is being planned, with the risk avoidance threshold $\phi$. If $\phi$ is exceeded, the cell is not entered.

\subsubsection{3D-DSP: Route composition}
\label{sec:compos}
A vehicle moving at 3km/h in the area of study is  faster than the current in any cell through which it travels\footnote{The fastest current in our region of the Southern ocean is the Antarctic Circumpolar Current (ACC) which, at its fastest, travels at 40-60cm$^s$ (about 2.1km/h)}. This allows routes to approximate direct paths between source and destination waypoints. The effect of the currents is seen in the selection of $yvals$ and the durations of the journeys, but paths are not distorted by currents pushing vehicles off-course. In contrast, vehicles travelling at 1km/h are pushed off-course by the faster currents (such as those in the Drake Passage) if no route to avoid the currents can be found. 

\begin{figure}
\centerline {\includegraphics[width=8cm]{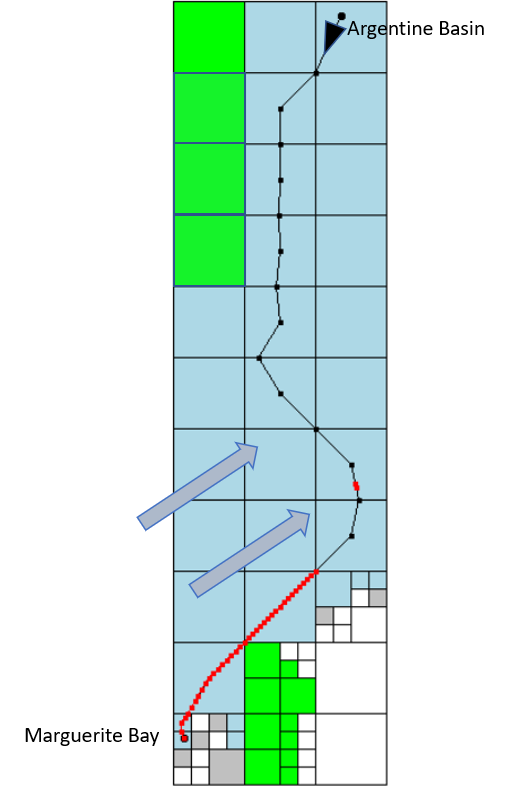}}
\caption{A 175-day journey planned in February 2013 for a 1km/h vehicle.}
\label{fig:ArgSea2MargBayFeb2013}
\end{figure}

Figure~\ref{fig:ArgSea2MargBayFeb2013} shows a projection of a route, from Argentine Basin to Marguerite Bay, planned using a 1km/h vehicle under the February 2013 monthly abstraction. The marked effect of the current (shown by the arrows) can be seen. Cells considered land are coloured green, while cells considered ice are white or grey, depending on their ice concentration in the planning month. The blue cells are open water.

This route was planned horizontally during the path-book construction, but is actually of almost 6 months duration. By the time the vehicle reaches the Northern Peninsula, the remainder of the proposed route is infeasible because it is in ice (the red points on the route shown where ice is encountered when the route is simulated starting in February 2013). 

Our path composition method removes infeasible sections of routes and replaces them with segments planned in months in which they can be completed without running into ice. A composed route consists of segments from the path-book, each planned in the associated month, assembled by Dijkstra's algorithm to achieve the shortest overall journey-time between source and destination. 

The composition is performed using a graph, $G_M$, representing paths available at different months in the year. $G_M = < V_M,E_M >$ is constructed with vertices, $V_M$, being the set of pairs $\langle Waypoint,Month \rangle$, and edges, $E_M$, as follows. If there is a path from \textit{wp1} to \textit{wp2} in month $m$, of at most 30 days' duration, then $E_M$ contains edges from $\langle \textit{wp1},m \rangle$ to $\langle \textit{wp2},m \rangle$ and $\langle \textit{wp1},m \rangle$ to $\langle \textit{wp2},m+1 \rangle$. If \textit{wp1} is accessible in months $m$ and $m+1$ then $E_M$ contains the edge $\langle \textit{wp1},m \rangle$ to $\langle \textit{wp1},m+1 \rangle$. This edge represents waiting at $\textit{wp1}$ during month $m$ and ensures waiting is possible only at ice-free locations.

\begin{figure}
\centerline {\includegraphics[width=7.5cm]{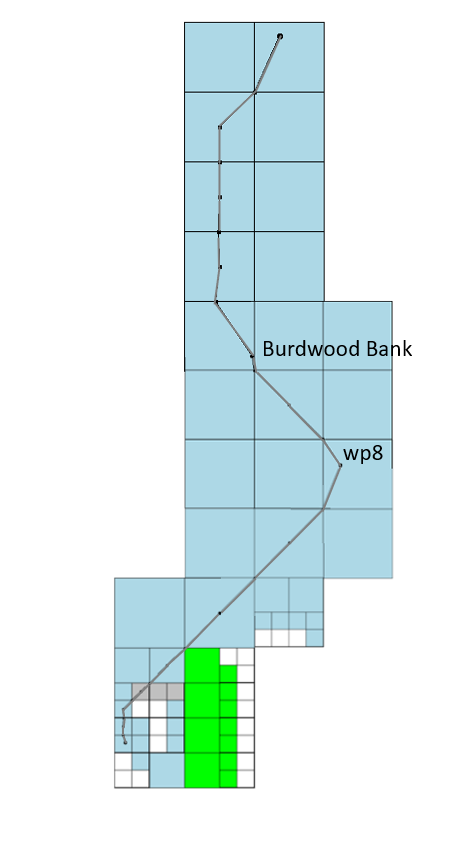}}
\caption{The composed journey for a 1km/h vehicle, from Argentine Basin to Marguerite Bay, leaving in December and arriving in May. The route consists of 5 subroutes, and the projection is made up from the 5 corresponding tiles. }
\label{fig:ArgSea2MargBayComposed}
\end{figure}
\begin{algorithm}
\bf{Algorithm: Compose($init$,$goal$,$G_M$)}\\
\KwData{A year graph, $G_M$, a specified source waypoint $init$, and a specified destination waypoint $goal$.}
\KwResult{The shortest path from $init$ to $goal$ in $G_M$.}
\Begin{
\textit{paths} $\leftarrow \emptyset$\;
\For{i=Jan..Dec}
{$\mathcal{S} \leftarrow \langle init,i \rangle $\;
$\mathcal{D} \leftarrow \emptyset$\;
\For{j=Jan..Dec}
{$\mathcal{D} \leftarrow \mathcal{D} \cup \langle goal,j \rangle $\;}
\textit{paths} $\leftarrow$ \textit{paths} $\cup$ Dijkstra($G_M,\mathcal{S},\mathcal{D}$)\;}
return \textit{shortest path in paths}\;}
\caption{Long paths are composed by running Dijkstra's algorithm over the path-book to find the shortest sequence of path components leading from $init$ to $goal$.}\label{alg2}
\end{algorithm}
In the use of Dijkstra's algorithm in Algorithm~\ref{alg2}, we track the total days travelled and prevent use of an edge {\em within} the same month once the arrival day extends into the next month.

The composed version of the route shown in Figure~\ref{fig:ArgSea2MargBayFeb2013} is given in Figure~\ref{fig:ArgSea2MargBayComposed}. The ice shown in the final segment is that of the May abstraction.

While the individual sections in the path-book minimise travel-time, the composed route may not be globally optimal as it is forced to pass between waypoints visited in the path-book components. This can be seen in Figure~\ref{fig:ArgSea2MargBayComposed}. Its structure is close to that of the optimal path in Figure~\ref{fig:ArgSea2MargBayFeb2013} but, because it is composed of 5 short paths, it visits a number of points en route (including Burdwood Bank and wp8). A detailed calendar format of this path can be found in section~\ref{sec:composedroutes}.

It may be necessary for the vehicle to wait en route, to allow iced-up regions to become accessible. The value of such a journey can be increased by electing to wait at intermediate waypoints of scientific interest or where the vehicle is safe from environmental impacts.  We provide an arbitrarily chosen collection of anonymous waypoints (identified as wps 1-34 in Appendix B) to serve this purpose, shown in Figure~\ref{fig:mapofwaypo}. We leave identification of scientifically useful and safe way-stations for future work.

\section{Evaluation of the Route-planning Method}
\label{sec:exps}
Our experiments focus on waypoint accessibility, route efficiency and the degree of exposure to risk implied by routes obtained under the different combinations of homogeneity configuration, risk avoidance and risk exposure settings. We contrast two speeds of vehicle: 1km/h, a speed typical of the commercially available gliders commonly used for ocean observation, and 3km/h, a speed more commonly associated with powered AUVs, and plausibly achievable by next-generation gliders. In all of the results presented we focus on 1-month routes, since these are the basic elements of all of our journeys. 

The number of waypoints accessible, and hence the number of journeys possible, in each year depends on the homogeneity configuration used and how risk-averse the route-planner is in deciding whether or not a cell is safe to enter. We run experiments with the Weak and Strong homogeneity configurations described in section~\ref{sec:nonhomog}. For each year, and for each combination of vehicle speed, risk-avoidance threshold, and homogeneity configuration, we generate a path-book consisting of twelve sets (one for each month) of optimal routes between all pairs of waypoints. 

We compare two risk avoidance thresholds for planning, $\phi = 5\%$ and $\phi = 8\%$. Both are used by all three homogeneity configurations. The impact of the threshold on waypoint accessibility is that the lower the value of $\phi$ the more risk-averse the route-planner is and the fewer routes are available in any year.

To evaluate the routes planned in a given configuration and year, $y_i$, we simulate running the routes against the {\em daily} ice concentrations recorded in the B-SOSE data set for a selected year $y_j$ ($y_i$ and $y_j$ may be the same or different). The daily ice data is taken as the ground truth against which we evaluate the impact of the monthly abstraction. We use a fixed threshold of $\psi = 12\%$ for determining risk exposure during simulation. 

\subsection{Risk Exposure}
\begin{figure}
         \centering
         \includegraphics[width=5cm]{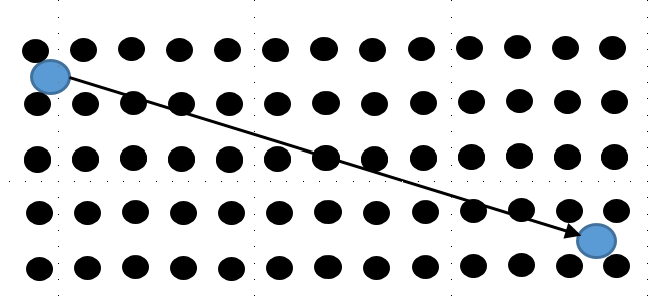}
         \caption{Data points used in the evaluation of a journey leg -- B-SOSE data at sixth of a degree resolution.}
         \label{fig:evalgrid}
\end{figure}

To determine the extent to which the route is exposed to risk, we consider each leg of the journey, where a leg is a constant bearing course from point to point, either from the centre of a cell to a crossing point, or a crossing point to a cell centre. Where a leg would start and end on different days (based on the duration of the journey up to the start point), the leg is subdivided to align with the day boundaries. In this way, we can associate each leg with a single day of ice conditions. The specific subset of B-SOSE data points that lie inside the rectangle (aligned with the latitude-longitude grid) defined by the end points of the leg (rounded to include points around the waypoints), as illustrated in Figure~\ref{fig:evalgrid}, is then averaged to give the average ice conditions encountered on that leg. 

As the vehicle traverses a leg in simulation, it either encounters ice above the $\psi = 12\%$ threshold or it does not. If it does, the duration of the leg is added to the total number of hours of risk encountered so far. At the end of the journey, the total number of risk hours is divided by 24 to give the number of {\em risk days} encountered on the journey. The number of risk days encountered at the $\psi = 12\%$ level is taken as a measure of the risk exposure experienced by the vehicle on the journey.

While the monthly abstraction captures how the ice extent changes between months, it obscures detail about how the ice extent changes {\em within} months. For example, the Weddell Sea starts January with extensive ice cover, which quickly collapses into February. The ice extent then builds steeply in March and April. A monthly abstraction of the sea ice in a cell into a mean value obscures these patterns.  

To explore how the ice growth and retreat processes affect exposure to risk, we count the number of risk days that occur within the month when travel was planned (the first 30 days of simulation). We separate our analysis into two parts: intra-year comparisons and inter-year comparisons. The intra-year comparisons consider routes planned under the monthly abstractions of a given year and simulated against the daily ice of that same year. The intra-year comparisons provide insights into the information content of the monthly abstraction itself. The inter-year comparisons consider routes planned under the monthly abstractions of each of the 5 years, simulated against the daily ice encountered in each of the other years. These comparisons show how robust the monthly abstraction is to annual variability.

Risk exposure is a consequence of the monthly abstraction, and can therefore never be avoided in planning, because some abstraction must always be made. Any risk exposure not accounted for by the monthly abstraction is avoided by using the composition method when a route of duration greater than 30 days is needed. As an optimisation, routes that are not affected by ice can be taken straight from the path-book.

\begin{figure*}
\centerline {\includegraphics[width=17cm]{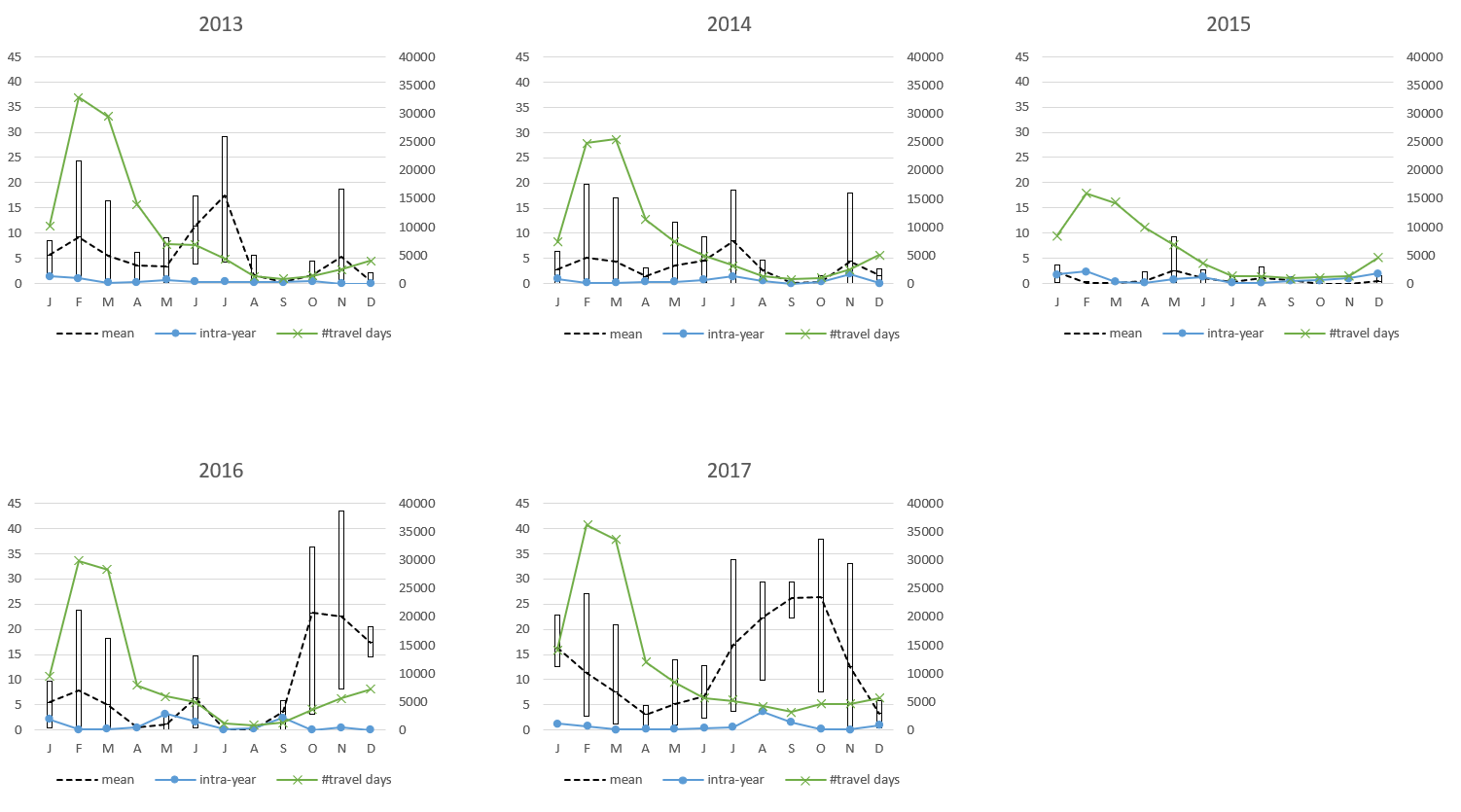}}
\caption{Risk exposure, showing routes generated in the year specified in the label, simulated in all 5 years. Plans are constructed using the 3km/h vehicle in Strong using $\phi = 8\%$. In each plot, the blue line is the intra-year percentage of travel days that are risk days. The black dotted line shows the mean percentage of risk days over all other years, encountered across all routes in each month (the inter-year comparison). The minimum and maximum bounds on risk exposure in each month are also shown. The green line indicates the number of travel days planned in each month.}
\label{fig:earlyviolsallyearsnew3km}
\end{figure*}

The details of the risk exposure encountered by vehicles travelling at 1km/h and 3km/h, using the Strong homogeneity configuration and the $\phi = 8$\% risk-avoidance threshold,  are given in Appendix B. The plots in Figures~\ref{fig:earlyviolsallyearsnew1km} and~\ref{fig:earlyviolsallyearsnew3km} represent the results of planning routes in the identified year and simulating them in all 5 years. 

\begin{figure*}
\centerline {\includegraphics[width=17cm]{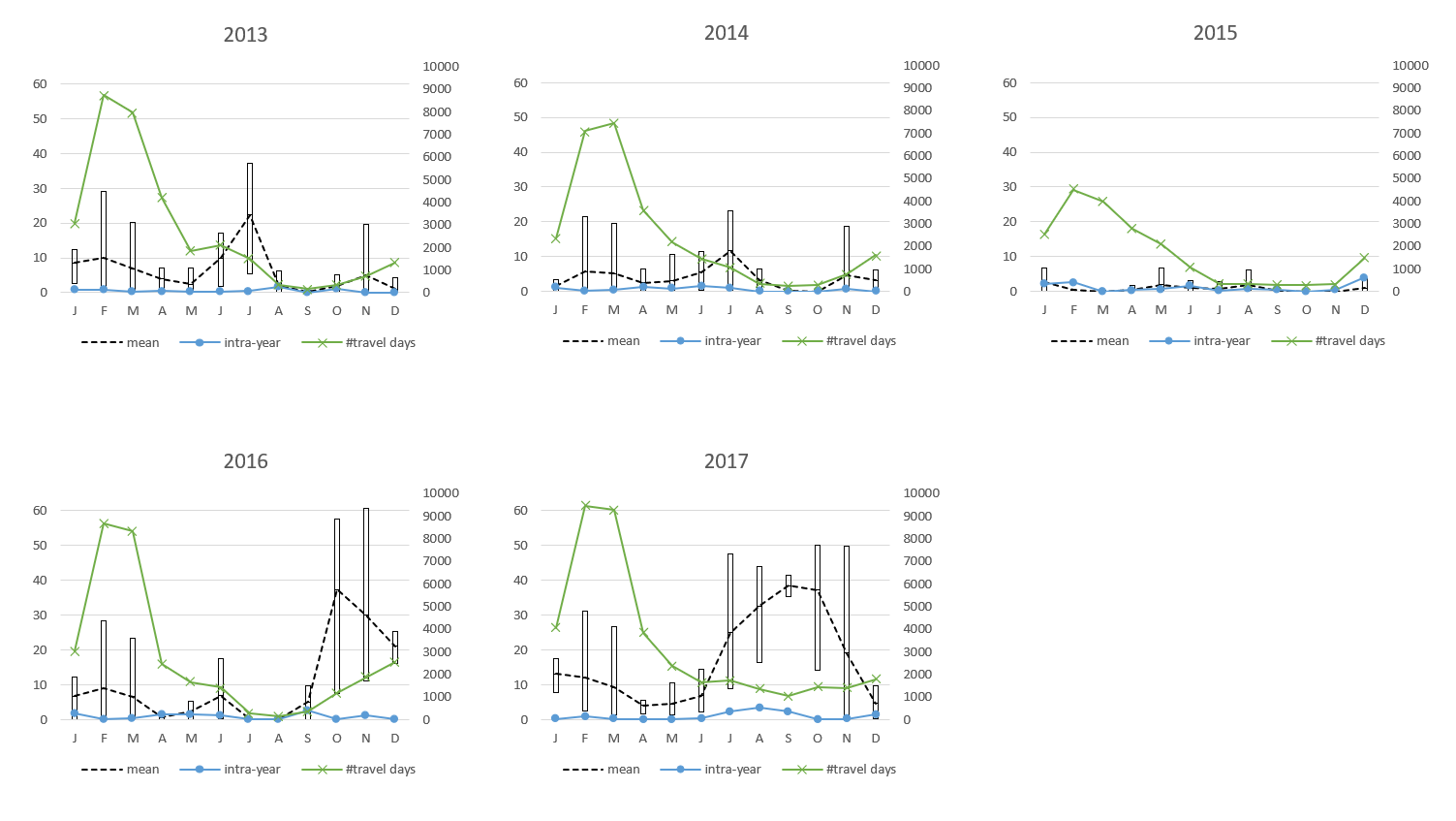}}
\caption{Risk exposure, showing routes generated in the year specified in the label, simulated in all 5 years. Plans are constructed using the 1km/h vehicle in Strong using $\phi = 8\%$. In each plot, the blue line is the intra-year percentage of travel days that are risk days. The black dotted line shows the mean percentage of risk days over all other years, encountered across all routes in each month (the inter-year comparison). The minimum and maximum bounds on risk exposure in each month are also shown. The green line indicates the number of travel days planned in each month.}
\label{fig:earlyviolsallyearsnew1km}
\end{figure*}

\begin{figure}
\centerline {\includegraphics[width=8cm]{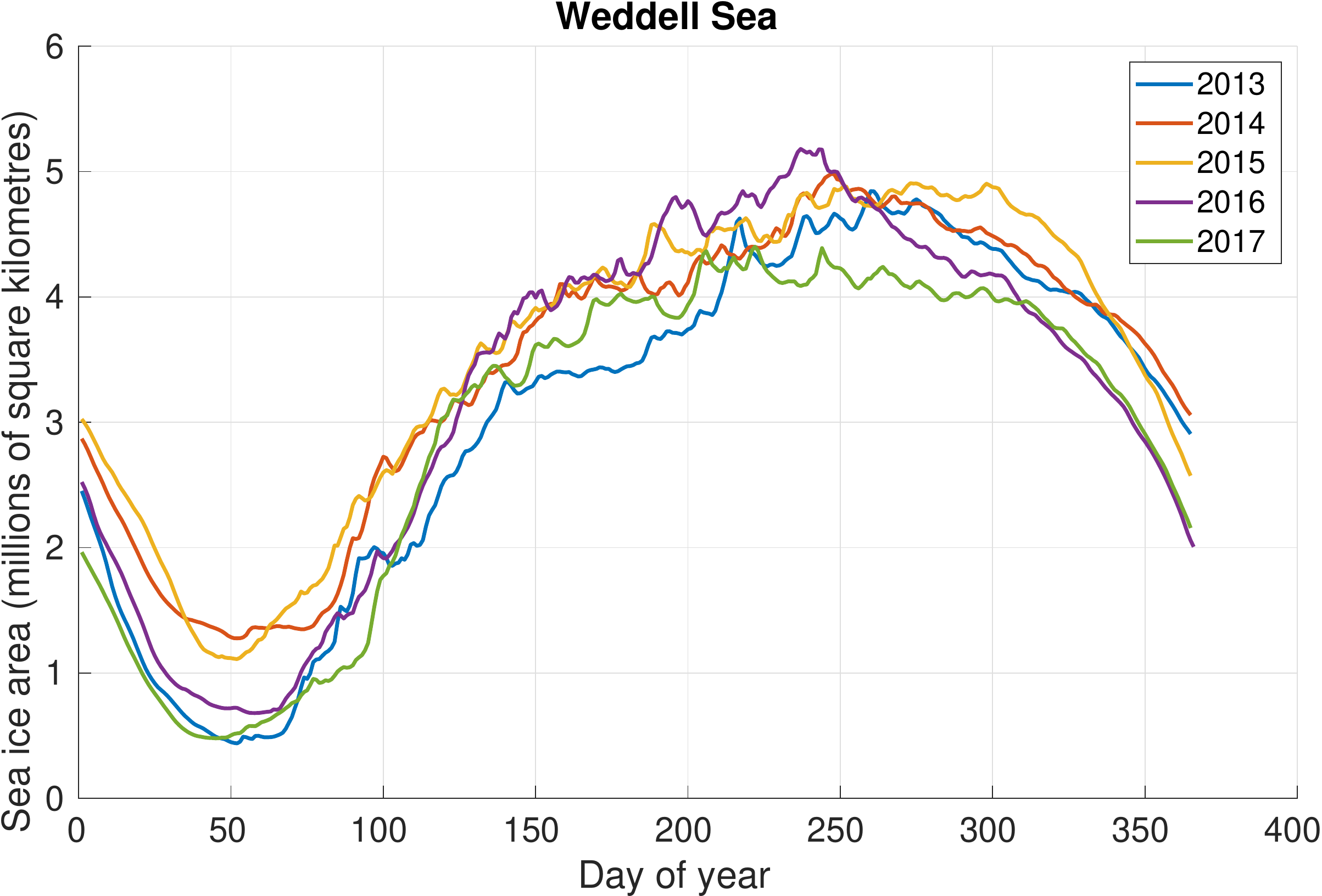}}
\caption{The ice extent in the Weddell Sea region in the 5 years 2013-2017. Data from the B-SOSE reanalysis product.}
\label{fig:weddellsea}
\end{figure}
It is interesting to compare the risk exposure results, shown in Appendix B, with the ice extent in the Weddell Sea region shown in Figure~\ref{fig:weddellsea}. Considering, for example, Figure~\ref{fig:earlyviolsallyearsnew3km}, the black lines in the plots quite faithfully follow the patterns of the corresponding years in Figure~\ref{fig:weddellsea}. In the months in which the ice in the planning year is less extensive than in other years, the mean risk exposure is greater, while when the ice is more extensive than in other years the mean risk exposure is smaller.  This is because, if route-planning is done under the pessimistic assumption that the ice extent is greater than in other years, the risk exposure encountered in simulation tends to be smaller than if an optimistic assumption is made. As an example, consider the plot for 2016 in Figure~\ref{fig:earlyviolsallyearsnew3km}, against the 2016 line in Figure~\ref{fig:weddellsea}. The 2016 line in Figure~\ref{fig:weddellsea} tracks the minimum sea ice area for the first 90 days of the year, and the means (and corresponding min and max bounds) seen in the 2016 plot of the figure are high over the first 3 months. The 2016 line then tracks the maximum ice until day 250, and the mean risk exposure in the autumn and winter months, shown in the plot, is correspondingly low over this period. After day 250, the 2016 line descends rapidly to track the minimum ice again over the final hundred days of the year, and the corresponding mean risk exposure is high in the final 3 months of the plot. The ranges of the min-max bounds in each month are largest when the gradient of the line is steepest, because the average over a rapidly changing month is less informative. A similar pattern to that seen in 2016 can be observed in all of the other plots of the figure. In particular, it can be noted that 2015 is essentially the maximal ice year according to Figure~\ref{fig:weddellsea}, and the plot for 2015 shows very low mean risk throughout (as well as very low waypoint accessibility). The plots for 2013 and 2014 both have low mean risk exposures in the last quarter of the year, and Figure~\ref{fig:weddellsea} shows that they both tracked the maximum sea ice area across all years in those months.

\subsection{Accessibility and Efficiency}
\begin{figure*}
         \centering
         \includegraphics[width=12cm]{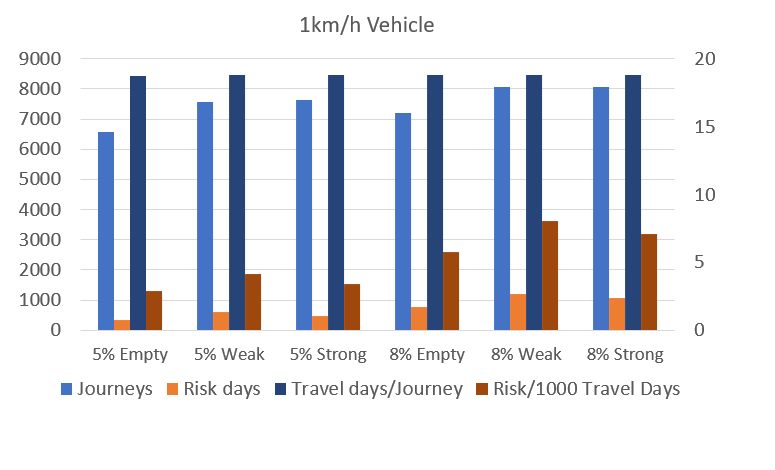}
         \caption{Waypoint accessibility, journey efficiency and daily risk using a 1km/h vehicle.}
         \label{fig:wpacc1km}
\end{figure*}

\begin{figure*}
         \centering
         \includegraphics[width=12cm]{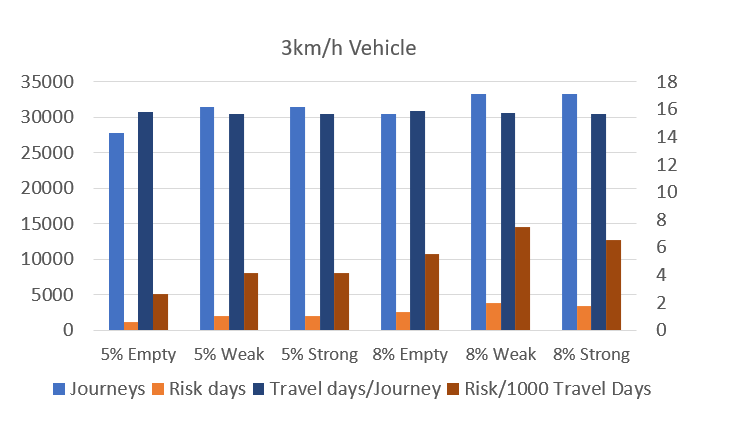}
         \caption{Waypoint accessibility, journey efficiency and daily risk using a 1km/h vehicle.}
         \label{fig:wpacc3km}
\end{figure*}

Figures~\ref{fig:wpacc1km} and~\ref{fig:wpacc3km} show how waypoint accessibility, route efficiency and risk exposure are affected by the choice of homogeneity configuration and risk avoidance threshold. 

In both figures, the numbers of journeys and risk days are read against the left-hand axis, and the two quotients are read against the right-hand axis. The figures presented are totals across all 5 years. The first 3 sets of columns relate to $\phi = 5$\%, the second set relates to $\phi = 8$\%. The number of routes available is taken as a measure of waypoint accessibility, while the number of travel days per journey is a measure of route efficiency. The number of intra-year risk days encountered is used as a measure of the intrinsic safety of the route-planning method, while the (scaled) risk/day measurement quantifies how intrinsically risky typical days of travel are. In Figure~\ref{fig:wpacc1km}, showing the results for the 1km/h vehicle, it can be seen that there are never more than 8000 routes available in any of the combinations of parameters considered. The number of routes available increases with further splitting. The average duration of a within-month journey is 18 days when a 1km/h vehicle is used. Risk exposure is always lower when $\phi = 5$\% is used than when $\phi = 8$\% is used, and the daily risk is also lower in the $\phi = 5$\%. However, interestingly, in both risk avoidance levels, daily risk drops off slightly in the Strong homogeneity configuration, indicating that increased splitting allows the ice to be more effectively avoided on some routes, without loss of accessibility or efficiency. This holds for both risk avoidance thresholds, indicating that a strong homogeneity condition is better than a weak one and, since there are more routes available under $\phi = 8$\% than under $\phi = 5$\%, a less conservative risk-avoidance threshold is also better. 

The same picture is seen in Figure~\ref{fig:wpacc3km}, except that -- predictably --  the number of routes available (and hence waypoint accessibility) is much greater when a faster vehicle is used. The average duration of a within-month journey is 16 days for the 3km/h vehicle, but the 3km/h vehicle covers about 3 times the distance covered by the 1km/h vehicle in the same period of time. The amount of risk exposure encountered is slightly higher for the 3km/h vehicle, but this is because a faster vehicle covers more ground and encounters a greater variety of conditions.

\subsection{Composed Routes}
\label{sec:composedroutes}
Routes of many months duration, covering thousands of kilometers through areas characterised by changing seasonal ice, can be constructed using the composition method. 

\begin{figure}
\centerline {\includegraphics[width=4cm]{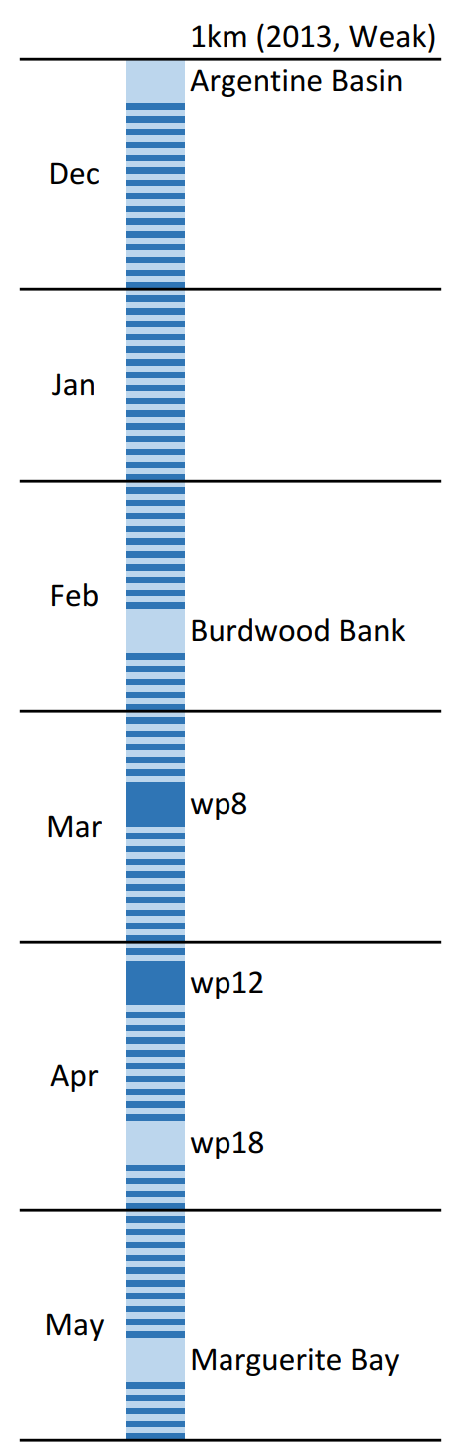}}
\caption{Slow vehicle composed path from Argentine Basin to Marguerite Bay, taking nearly 6 months.}
\label{fig:slowmover}
\end{figure}

\begin{figure}
\centerline {\includegraphics[width=4cm]{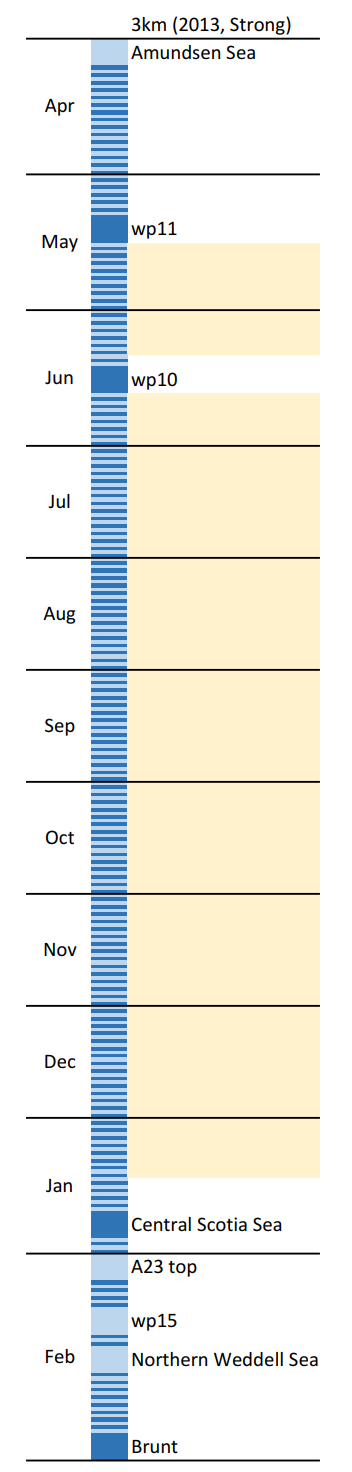}}
\caption{Fast vehicle composed path from Amundsen Sea to Brunt, taking 11 months, including 6 months waiting at wp10 (waiting periods shown highlighted in yellow).}
\label{fig:fastmover}
\end{figure}

Tables~\ref{tab:3kmcomp} and~\ref{tab:1kmcomp} show a selection of journeys constructed using our composition method, using the 3km/h and 1km/h vehicles. Given our models, some of the destination waypoints (such as Brunt and Northern Weddell Sea) can only be entered at very limited times of year. For example, Brunt is only accessible in February when $\phi$ is as low as 8\% (a less conservative risk-avoidance threshold would increase its availability). Even using a 3km/h vehicle, the best journey to Brunt from Amundsen Sea ($\sim$5,000km) must start in April and spend 7 months waiting en route to get there in time. Waiting is sometimes unavoidable on such long journeys. The journeys shown are the shortest of all possible paths between the specified waypoints in the given year. Figure~\ref{fig:fastmover}, in section~\ref{sec:composedroutes}, is a calendar format visualisation of the Amundsen Sea to Brunt journey using the 3km/h vehicle. The horizontal lines in the blue column mark the days of travel, with the visited waypoints marked on the right (their coordinates can be found in Appendix B), and the months of travel on the left. The yellow periods are the waiting times along the journey. Figure~\ref{fig:slowmover} is a calendar format visualisation of the composed version of the journey, using a 1km/h vehicle, from Argentine Sea to Marguerite Bay, shown in  Figure~\ref{fig:ArgSea2MargBayFeb2013}. 

\begin{table*}[ht]
\centering
\begin{tabular}{lcccc}
Route & Year & Starting & Waiting & Duration (days)\\
\hline
Amundsen Sea$\rightarrow$Brunt & 2017 & April & 7 & 328.7\\
Palmer$\rightarrow$South Georgia & 2015 & January & 0 & 26.8\\
South Sandwich Trench$\rightarrow$Elephant Island  & 2014 & April & 0 & 27.5\\
South Georgia$\rightarrow$SR4bottom &  2017 & February & 0 & 24.5\\
Maud Rise$\rightarrow$Marguerite Bay & 2013 & February & 0 & 52.4\\
Amundsen Sea$\rightarrow$Northern Weddell Sea & 2017 & February & 0 & 63.5\\
\hline

\end{tabular}
\caption{Examples of composed routes using the 3km/h vehicle.}
\label{tab:3kmcomp}
\end{table*}

\begin{table*}[ht]
\centering
\begin{tabular}{lcccc}
Route & Year & Starting & Waiting (months) & Duration (days)\\
\hline
Amundsen Sea$\rightarrow$Brunt & 2017 & - & - & -\\
Palmer$\rightarrow$South Georgia & 2015 & February & 0 & 67.4\\
South Sandwich Trench$\rightarrow$Elephant Island  & 2014 & February & 0 & 94\\
South Georgia$\rightarrow$SR4bottom &  2017 & January & 0 & 82.1\\
Maud Rise$\rightarrow$Marguerite Bay & 2013 & February & 5 & 403\\
Amundsen Sea$\rightarrow$Northern Weddell Sea & 2017 & April & 6 & 346.8\\
\hline

\end{tabular}
\caption{Examples of composed routes using the 1km/h vehicle.}
\label{tab:1kmcomp}
\end{table*}


\section{Summary}

We have developed a long-distance route-planning capability, 3D-DSP, for underwater vehicles operating in icy regions. 
Our results show that:
\begin{enumerate}
\item Our combination of Dijkstra's algorithm, Newton's method and monthly grid refinement helps to overcome some of the limitations of existing grid-based discretisations and wave propagation approaches without significant computational expense.
\item Using different combinations of risk avoidance threshold, $\phi$, and homogeneity conditions, we can plan more or less efficient and risk-averse journeys. Increased grid refinement can result in lower risk exposure because of greater awareness of the locations of possible dangers. 
\item We can plan journeys of thousands of kilometers taking into account the surface currents and the monthly behaviour of the sea ice in our segment of the Southern Ocean.
\end{enumerate}

We based this work on an historical dataset comprising 5 years of Southern Ocean State Estimate reanalysis data. These 5 years clearly vary in terms of sea ice extent in our region of study, as seen in Figure~\ref{fig:weddellsea}. Choosing a maximal ice year for planning results in very little inter-year risk exposure, but also very low waypoint accessibility. By contrast,  choosing a minimal year for planning gives high waypoint accessibility at the cost of relatively high risk exposure. A trade-off must be made to meet operational needs.  

The product of our work is a long-distance route planning capability, 3D-DSP, analogous to that of the Tethys Long-Range AUV, for polar ocean conditions. Our approach does not yet consider icebergs, as distinct from sea ice. This will be addressed in future work. In addition, there are several other ideas that remain to be explored. As mentioned, we are now developing a full risk model to replace the risk exposure measurements. This will allow a cost function that combines travel time with risk minimisation. Future work will also consider other homogeneity conditions, for example: conditions based on characteristics such as land, current, bathymetry, icebergs and other features including the temporal variability of sea ice within months. Furthermore, as our methods are fully generic, long-distance route-planning for different types of vehicles, constrained by different operational limitations and responsive to different environmental constraints, can be addressed.

\section*{Acknowledgments}
The work described in this paper has been supported by a number of grants. Fox was supported by a Royal Society Industry Fellowship, INF\textbackslash R1\textbackslash 201041, Brearley by NERC Independent Research Fellowship NE/L011666/1 and also through NERC National Capability programmes NE/R016038/1 and NE/V013254/1, and Jones by a UKRI Future Leaders Fellowship (reference MR/T020822/1).

\section*{Appendix A: Technical Result}
The following is a result that demonstrates that the straight line path between the centres of adjacent squares will always cross the shared edge segment. 

\begin{theorem}
\label{theorem}
Given two squares, $A$ and $B$, such that one edge of the smaller square (without loss of generality, $B$) lies entirely within one edge of the other square, the  line connecting the centres of the two squares crosses the edge of $B$ that is part of the edge of $A$.
\end{theorem}

\begin{figure}
    \centering
    \includegraphics{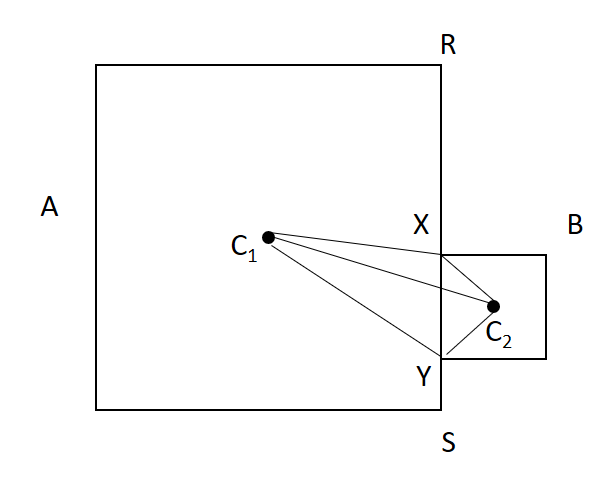}
    \caption{Two adjacent squares, $A$ and $B$, with $B$ sharing its entire edge $XY$ with edge $RS$ of $A$.}
    \label{fig:squares}
\end{figure}

\noindent
{\bf Proof:} Without loss of generality, take the arrangement of the squares to be as shown in Figure~\ref{fig:squares}, with the line $C_1C_2$ connecting the centres of the two squares. The angle $\angle XC_2Y$ is $90\degree$. The angle $\angle SC_1Y$ satisfies $0\degree \leq \angle SC_1Y \leq 90\degree$ because $Y$ lies within $RS$. By complementary angles, $\angle C_1YX = 45\degree + \angle SC_1Y$ and, therefore, $45\degree \leq \angle C_1YX \leq 135\degree$. The angle $\angle C_2YX$ is $45\degree$, so $90\degree \leq \angle C_2YC_1 \leq 180\degree$. By a similar argument, $90\degree \leq \angle C_1XC_2 \leq 180\degree$.
Since $XY$ is contained within $RS$, the angle $\angle XC_1Y \leq 90\degree$. Thus, $XC_1YC_2$ forms a convex quadrilateral and, therefore, its diagonals intersect inside it. The intersection of the diagonals is exactly the crossing point of the line $C_1C_2$ at the edge of $A$ and $B$, which completes the proof.

Note that the claim remains true, trivially, when $A$ and $B$ are of equal size and that the condition that $B$ is not degenerate (ie $X \not = Y$) ensures that the quadrilateral $XC_1YC_2$ is also not degenerate.

\section*{Appendix B: Waypoints and their Coordinates}

We use 29 named waypoints of scientific interest (Table~\ref{table:the_waypoints}), and an additional 34 waypoints Table~\ref{table:anon_waypoints}) scattered around and between these waypoints to serve as via points and way stations for use on long traverses. 

\begin{table}[ht]
\centering
\begin{tabular}{l *{2}{c}}
\hline
Name & Latitude & Longitude\\
\hline
1. Falklands	& -52$^{\circ}$38' 05''	& -59$^{\circ}$52' 48''\\
2. South Georgia	& -54$^{\circ}$52' 45''	& -37$^{\circ}$15' 51''\\
3. South Sandwich Trench	& -56$^{\circ}$24' 41''	& -25$^{\circ}$03' 31''\\
4. Northern Peninsula	& -63$^{\circ}$01' 53''	& -50$^{\circ}$56' 24''\\
5. Maud Rise	& -66$^{\circ}$04' 58''	& -2$^{\circ}$49' 39''\\
6. Amundsen Sea	& -69$^{\circ}$19' 23''	& -129$^{\circ}$41' 27''\\
7. Central Weddell Sea	& -71$^{\circ}$05' 05''	& -44$^{\circ}$56' 43''\\
8. South Orkney Islands	& -60$^{\circ}$29' 10''	& -45$^{\circ}$18' 17''\\
9. Central Scotia Sea	& -58$^{\circ}$21' 29''	& -44$^{\circ}$13' 58''\\
10. Filchner Trough Overflow & -76$^{\circ}$59' 34''	& -35$^{\circ}$56' 49''\\
11. Filchner-Ronne Ice Shelf & -77$^{\circ}$58' 05''	& -41$^{\circ}$29' 56''\\
12. Argentine Basin	& -43$^{\circ}$00' 00''	& -58$^{\circ}$08' 02''\\
13. Bellingshausen Sea	& -70$^{\circ}$54' 12''	& -82$^{\circ}$49' 03''\\
14. Marguerite Bay	& -68$^{\circ}$23' 21''	& -69$^{\circ}$14' 37''\\
15. Palmer	& -64$^{\circ}$40' 43''	& -67$^{\circ}$07' 59''\\
16. Brunt	& -74$^{\circ}$02' 30''	& -28$^{\circ}$17' 05''\\
17. Northern Weddell Sea	& -63$^{\circ}$26' 55''	& -34$^{\circ}$12' 07''\\
18. NorthWest Georgia Rise	& -52$^{\circ}$21' 25''	& -36$^{\circ}$56' 16''\\
19. Shag Rocks Passage	& -52$^{\circ}$12' 50''	& -48$^{\circ}$09' 20''\\
20. Elephant Island	& -60$^{\circ}$32' 50''	& -55$^{\circ}$10' 53''\\
21. Burdwood Bank	& -54$^{\circ}$29' 07''	& -60$^{\circ}$08' 20''\\
22. A23 Transect (bottom)	& -62$^{\circ}$26' 3.07''	& -33$^{\circ}$15' 39.28''\\
23. A23 Transect (top)	& -58$^{\circ}$1' 13.83''	& -38$^{\circ}$21' 30.47''\\
24. SR4 Transect (top)	& -61$^{\circ}$34' 57.72''	& -49$^{\circ}$45' 21.68''\\
25. SR4 Transect (bottom)	& -68$^{\circ}$24' 47.72''	& -29$^{\circ}$18' 47.23''\\
26. Orkney Passage Entry	& -60$^{\circ}$57' 5.18''	& -28$^{\circ}$34' 43.48''\\
27. Orkney Passage Exit	& -60$^{\circ}$41' 40.18''	& -39$^{\circ}$10' 34.45''\\
28. Weddell Sea 1 & -70$^{\circ}$46' 05''	& -52$^{\circ}$58' 29''\\
29. Weddell Sea 2 & -66$^{\circ}$46' 27''	& -57$^{\circ}$15' 11''\\
\hline
    \end{tabular}
    \caption{A collection of waypoints of scientific interest located in the region of study}
    \label{table:the_waypoints}
\end{table}

Table~\ref{table:anon_waypoints} shows the 34 anonymous waypoints, and their coordinates.
\begin{table}[ht]
\centering
\begin{tabular}{l *{2}{c}}
\hline
Identifier & Latitude & Longitude\\
\hline
wp1	& -53$^{\circ}$ 45' 0''	& -32$^{\circ}$ 30' 0''\\
wp2	& -55$^{\circ}$ 56' 0''	& -28$^{\circ}$ 7' 0''\\
wp3	& -56$^{\circ}$ 15' 0''	& -72$^{\circ}$ 30' 0''\\
wp4	& -56$^{\circ}$ 15' 0''	& -37$^{\circ}$ 30' 0''\\
wp5	& -56$^{\circ}$ 18' 0''	& -55$^{\circ}$ 0' 0''\\
wp6	& -57$^{\circ}$ 29' 0''	& -45$^{\circ}$ 0' 0''\\
wp7	& -58$^{\circ}$ 26' 0''	& -56$^{\circ}$ 52' 0''\\
wp8	& -58$^{\circ}$ 26' 0''	& -53$^{\circ}$ 45' 0''\\
wp9	& -58$^{\circ}$ 26' 0''	& -53$^{\circ}$ 7' 0''\\
wp10	& -59$^{\circ}$ 59' 0''	& -57$^{\circ}$ 22' 0''\\
wp11	& -61$^{\circ}$ 14' 0''	& -62$^{\circ}$ 29' 0''\\
wp12	& -61$^{\circ}$ 14' 0''	& -57$^{\circ}$ 29' 0''\\
wp13	& -61$^{\circ}$ 14' 0''	& -52$^{\circ}$ 30' 0''\\
wp14	& -61$^{\circ}$ 14' 0''	& -47$^{\circ}$ 30' 0''\\
wp15	& -61$^{\circ}$ 33' 0''	& -36$^{\circ}$ 15' 0''\\
wp16	& -62$^{\circ}$ 29' 0''	& -84$^{\circ}$ 59' 0''\\
wp17	& -63$^{\circ}$ 26' 0''	& -37$^{\circ}$ 30' 0''\\
wp18	& -63$^{\circ}$ 45' 0''	& -62$^{\circ}$ 29' 0''\\
wp19	& -63$^{\circ}$ 45' 0''	& -47$^{\circ}$ 30' 0''\\
wp20	& -63$^{\circ}$ 45' 0''	& -32$^{\circ}$ 30' 0''\\
wp21	& -63$^{\circ}$ 45' 0''	& -22$^{\circ}$ 30' 0''\\
wp22	& -63$^{\circ}$ 45' 0''	& -29$^{\circ}$ 59' 0''\\
wp23	& -65$^{\circ}$ 0' 0''	& -65$^{\circ}$ 0' 0''\\
wp24	& -65$^{\circ}$ 37' 0''	& -21$^{\circ}$ 15' 0''\\
wp25	& -66$^{\circ}$ 15' 0''	& -42$^{\circ}$ 30' 0''\\
wp26	& -67$^{\circ}$ 30' 0''	& -40$^{\circ}$ 0' 0''\\
wp27	& -71$^{\circ}$ 15' 0''	& -92$^{\circ}$ 30' 0''\\
wp28	& -71$^{\circ}$ 15' 0''	& -82$^{\circ}$ 29' 0''\\
wp29	& -71$^{\circ}$ 15' 0''	& -77$^{\circ}$ 30' 0''\\
wp30	& -66$^{\circ}$ 0' 0''	& -75$^{\circ}$ 0' 0''\\
wp31	& -67$^{\circ}$ 0' 0''	& -117$^{\circ}$ 0' 0''\\
wp32	& -67$^{\circ}$ 0' 0''	& -106$^{\circ}$ 0' 0''\\
wp33	& -67$^{\circ}$ 0' 0''	& -102$^{\circ}$ 0' 0''\\
wp34	& -71$^{\circ}$ 0' 0''	& -111$^{\circ}$ 0' 0''\\
\hline
    \end{tabular}
    \caption{A scattering of anonymous waypoints used as via points in the region of study.}
    \label{table:anon_waypoints}
\end{table}

\section*{Appendix C: Derivation of the Expression for $yval$ in the Refined Grid}
\begin{figure}
\centerline {\includegraphics[width=6cm]{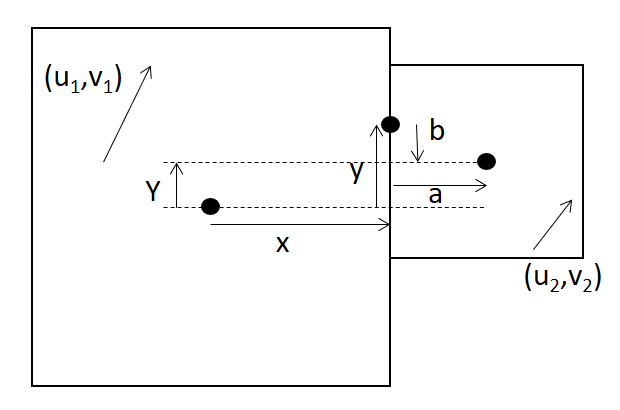}}
\caption{Finding the optimal $yval$ when crossing between different sized grid cells. }
\label{fig:yvalderiv}
\end{figure}

Here we provide the method for deriving $yval$ when crossing between cells of different sizes. The variable $y$ stands for the optimal crossing point value $yval$. Figure~\ref{fig:yvalderiv} provides the context for interpreting the following equations.

Figure~\ref{fig:yvalderiv} shows how the optimal $yval$ is shown in the case where the adjacent cells are of different sizes. The case in which the large cell is on the left is shown.
Figure~\ref{fig:yvalderiv} refers to the following quantities. $t_1$ and $t_2$ are the travel times in the left and right cells respectively. $(u_1,v_1)$ and $(u_2,v_2)$ are the current vector components in the left and right cells respectively. $Y$ is the vertical separation of the two centre points. $x$ is half of the width of the left cell, and $a$ is half of the width of the right cell. As noted above, $y$ is $yval$ itself. $b$ is the vertical distance travelled in the right hand cell. $s$ is the speed of the vehicle.

 The distances travelled in the two cells are defined as follows.
\[\begin{array}{ll}
d_1^2 = x^2 + y^2\\
d_2^2 = a^2 + b^2 = a^2 + (Y-y)^2\\
\end{array}\]

The following terms are helpful in simplifying the derivations below.
\[\begin{array}{ll}
C_1 = s^2 - u_1^2 - v_1^2\\
C_2 = s^2 - u_2^2 - v_2^2\\
D_1 = xu_1 + yv_1\\
D_2 = au_2 + (Y-y)v_2\\
X_1 = \sqrt{D_1^2 + C_1d_1^2}\\
X_2 = \sqrt{D_2^2 + C_2d_2^2}\\
\end{array}\]
Using these expressions, Equation~\ref{eqn:eqn6} can be written as follows for the two cells.
\begin{equation}
\label{t1a}
    t_1 = \frac{X_1 - D_1}{C_1}
\end{equation}
\begin{equation}
\label{t2a}
    t_2 = \frac{X_2 - D_2}{C_2}
\end{equation}
Differentiation of Equation~\ref{eqn:eqn5} with respect to $y$, in each of the two cells, yields:
\begin{equation}
\label{eqn:t1diff}
\frac{dt_1}{dy}(C_1t_1+D_1) = y-t_1v_1
\end{equation}
\begin{equation}
\label{eqn:t2diff}
\frac{dt_2}{dy}(C_2t_2+D_2) = y-Y+t_2v_2
\end{equation}
The minimum travel time is achieved when $\frac{dt_1}{dy} + \frac{dt_2}{dy} = 0$. Therefore, using equations~\ref{eqn:t1diff} and~\ref{eqn:t2diff} we require $F(y) = 0$ where:

\begin{equation}
F(y) = X_1(y - Y + \frac{v_2(X_2 - D_2)}{C_2}) + X_2(y - \frac{v_1(X_1 - D_1)}{C_1}) 
\end{equation}

Referring back to Equation~\ref{newton}, note that: \[F(y) = X_1X_2 (ttl'(y) + ttr'(y)) \]
 
The derivation in section~\ref{newtderiv} is a specialisation of the above, in which $Y = 0$ and $a = x$ (the case in which the adjacent cells are the same size).


\bibliographystyle{plain}
\bibliography{RPMesh}

\end{document}